\documentclass[lettersize,journal]{IEEEtran}
\usepackage{amsmath,amsfonts}
\usepackage{algorithmic}
\usepackage{algorithm}
\usepackage{array}
\usepackage[caption=false,font=normalsize,labelfont=sf,textfont=sf]{subfig}
\usepackage{textcomp}
\usepackage{stfloats}
\usepackage{url}
\usepackage{verbatim}
\usepackage{graphicx}
\usepackage{cite}
\usepackage{adjustbox}
\usepackage{booktabs}
\usepackage{multirow}
\usepackage{color}
\usepackage{hyperref}
\begin{document}

\title{MS-CustomNet: Controllable Multi-Subject Customization with Hierarchical Relational Semantics}

\author{Pengxiang Cai and Mengyang Li
\thanks{Pengxiang Cai is with the School of Information Science and Engineering, East China University of Science and Technology, Shanghai 200237, China.}
\thanks{Mengyang Li is with the Center for Applied Mathematics, Tianjin University, Tianjin 300072, China.}
\thanks{Corresponding author: Mengyang Li (e-mail: limengyang@tju.edu.cn).}}

\markboth{IEEE TRANSACTIONS ON HUMAN-MACHINE SYSTEMS,~Vol.~XX, No.~X, MONTH~2025}%
{Cai and Li: MS-CustomNet: Controllable Multi-Subject Customization with Hierarchical Relational Semantics}

\maketitle

\begin{abstract}
Diffusion-based text-to-image generation has advanced significantly, yet customizing scenes with multiple distinct subjects while maintaining fine-grained control over their interactions remains challenging. Existing methods often struggle to provide explicit user-defined control over the compositional structure and precise spatial relationships between subjects. To address this, we introduce MS-CustomNet, a novel framework for multi-subject customization. MS-CustomNet allows zero-shot integration of multiple user-provided objects and, crucially, empowers users to explicitly define these hierarchical arrangements and spatial placements within the generated image. Our approach ensures individual subject identity preservation while learning and enacting these user-specified inter-subject compositions. We also present the MSI dataset, derived from COCO, to facilitate training on such complex multi-subject compositions. MS-CustomNet offers enhanced, fine-grained control over multi-subject image generation. Our method achieves a DINO-I score of 0.61 for identity preservation and a YOLO-L score of 0.94 for positional control in multi-subject customization tasks, demonstrating its superior capability in generating high-fidelity images with precise, user-directed multi-subject compositions and spatial control.
\end{abstract}


\section{Introduction}

\IEEEPARstart{T}{he} field of generative artificial intelligence, powered by diffusion models \cite{ho2020denoising, rombach2022high, saharia2022photorealistic}, has revolutionized image synthesis from text. As this technology matures, the frontier of innovation is shifting from generating visually pleasing images to providing users with fine-grained, predictable control over the content. A critical aspect of this control is subject customization: integrating user-provided subjects into novel scenes while faithfully preserving their identity.

Initial subject customization efforts utilized optimization-based techniques \cite{ruiz2023dreambooth, gal2022image, kumari2023customdiffusion}, fine-tuning models or learning embeddings per concept. While often achieving high fidelity, these incur significant computational costs and can overfit with limited views. Encoder-based approaches \cite{li2023blip, wei2023elite, yuan2024customnet} overcome this by training an encoder to map reference images to conditional embeddings for zero-shot customization. Notably, CustomNet \cite{yuan2024customnet} advanced single-subject customization by incorporating explicit 3D viewpoint conditioning, leveraging insights from novel view synthesis \cite{liu2023zero}. This allowed control over the single subject's spatial orientation and placement, improving identity preservation and integration.

However, extending customization from a single subject to multiple distinct subjects within one scene—the focus of Multi-Subject CustomNet (MS-CustomNet)—introduces new challenges. Coherent generation requires preserving each subject's identity while accurately modeling their complex spatial relationships, semantic interactions, and consistent integration within a shared background, guided by a potentially intricate prompt. Naive application of single-subject techniques often leads to concept bleeding, identity loss, or failure to capture intended compositional semantics \cite{shi2024instantbooth, yang2024rpg}.

While methods like MS-Diffusion \cite{wang2024msd} achieve state-of-the-art fidelity in general multi-subject synthesis, we pursue a fundamentally different goal. MS-Diffusion excels at high-fidelity rendering guided by the implicit semantics of a text prompt. In contrast, MS-CustomNet is expressly designed for applications demanding explicit and reproducible control over the compositional structure (e.g., relative layering or occlusion) and precise spatial relationships. For a user needing to place object A specifically inside object B, the implicit nature of attention mechanisms in general-purpose models offers limited predictability. Our work addresses this gap by delivering this enhanced, deterministic control within an accessible and efficient framework.

To address these limitations in fine-grained multi-subject control, we propose MS-CustomNet, a novel framework for multi-subject customization. MS-CustomNet facilitates the zero-shot integration and controllable placement of multiple user-provided objects within a generated image. To combat concept loss, our approach employs a decoupled feature learning mechanism that preserves individual subject identities. To enable explicit compositional arrangement, we introduce a relational modeling component that learns inter-subject semantics directly from user-provided layout guidance. Recognizing the lack of suitable training data, we introduce the Multi-Subject Interaction (MSI) dataset. Constructed using COCO annotations \cite{lin2014coco}, MSI provides examples of multi-subject compositions, facilitating learning of complex inter-subject relationships.

MS-CustomNet offers an end-to-end framework for more generalized control of compositional structures and spatial relationships. The main contributions are:
\begin{enumerate}
\item A novel framework, MS-CustomNet, extending single-subject customization to the multi-subject domain, specifically designed for generating images with multiple customized subjects and offering explicit, user-defined control over their compositional structure.
\item Novel training strategies, namely Dual Stage Training and Curriculum Learning on Subject Quantity, designed to enhance model stability and efficacy in learning complex multi-subject compositions and identity preservation.
\item The introduction of the Multi-Subject Interaction dataset, derived from COCO, to facilitate training and evaluation of models focused on multi-subject compositional structure and spatial relationship.
\item Demonstration of MS-CustomNet’s efficacy in zero-shot multi-subject customization, achieving preserved subject identities, controllable placement, and semantically coherent scenes that adhere to user-defined compositional structures and inter-subject relationships, all within an efficient framework.
\end{enumerate}

\section{Related Work}

Generative modeling, particularly with diffusion models, has achieved remarkable progress in synthesizing high-fidelity images from text \cite{zhang2025scale, yang2024progressive, zhang2024diff, chen2024diffusion}. A significant focus within this field has been personalized image generation, enabling users to incorporate specific visual concepts into the creative process.

\subsection{Subject Customization in Diffusion Models}

Current subject customization techniques primarily fall into two categories: optimization-based and encoder-based methods.

Optimization-based methods, such as the pioneering DreamBooth \cite{ruiz2023dreambooth} and Textual Inversion \cite{gal2022image}, achieve high-fidelity subject representation by fine-tuning parts of the diffusion model or learning dedicated textual embeddings for the target concept using a small set of reference images. Subsequent methods like CustomDiffusion \cite{kumari2023customdiffusion}, DisenBooth \cite{chen2023disenbooth}, Cones \cite{liu2023cones}, and JEDI \cite{zeng2024jedi} have further explored optimizing specific network parameters or disentangling identity from context. While effective in preserving identity, these methods require per-subject optimization, which is time-consuming and can be prone to overfitting, especially with limited reference views \cite{yuan2024customnet}, making them less scalable for rapid or dynamic customization needs. Orthogonal fine-tuning methods \cite{po2024orthogonal} have also been proposed to mitigate catastrophic forgetting during customization.

To overcome the limitations of optimization, encoder-based approaches aim for zero-shot or few-shot customization by training an auxiliary encoder to map reference images into the conditioning space of a pre-trained diffusion model \cite{yang2023paint, song2023objectstitch, shi2024instantbooth}. These methods learn to extract visual concepts directly from images at inference time. For instance, BLIP-Diffusion \cite{li2023blip} leverages pre-trained vision-language models. ELITE \cite{wei2023elite} fuses deep and shallow features to enhance detail preservation. GLIGEN \cite{li2023gligen} incorporates spatial grounding information. However, early encoder-based methods often struggled with preserving fine-grained details or suffered from a `copy-paste' effect \cite{yuan2024customnet}, lacking sufficient variation or harmonious integration with the background. CustomNet \cite{yuan2024customnet}, a key reference point for our work, significantly improved single-subject customization by explicitly incorporating viewpoint control, leveraging techniques from novel view synthesis \cite{liu2023zero} to generate varied yet identity-preserving results with better scene harmony. Scalable approaches like SVDiff \cite{peebles2023scalable} and methods for video customization \cite{wei2024dreamvideo, guo2023animatediff} also fall under this paradigm.

\subsection{Multi-Subject Generation and Control}

While single-subject customization has advanced considerably, extending these techniques to handle multiple distinct subjects within a single generated image presents unique challenges. Directly applying single-subject methods often leads to concept bleeding, identity loss, or confused interactions between subjects \cite{yang2024rpg, shi2024instantbooth, ding2024freecustom}.

Several recent works have started to address multi-concept or multi-subject generation. Some methods focus on compositional generation by manipulating attention maps or latent codes \cite{blattmann2023align, ke2024repurposing}. Others propose specialized training strategies or architectures. For example, some approaches denoise different subjects or regions separately before fusion \cite{ding2024freecustom, chakrabarty2024lomoe, dong2024cidm, yang2024rpg}, aiming to preserve individual identities. RelationBooth \cite{shi2024relationbooth} explicitly models subject relationships and poses using keypoint guidance and separate encoders. InstantBooth \cite{shi2024instantbooth} incorporates image patch features into the diffusion U-Net. However, achieving robust identity preservation for all specified subjects while simultaneously modeling their complex interactions and ensuring accurate correspondence with textual prompts remains an open challenge. Many existing multi-subject approaches still struggle with semantic entanglement or fail to accurately map textual descriptions to the correct visual instances \cite{shi2024relationbooth}.

Our MS-CustomNet method directly tackles this multi-subject customization gap. Building on the insights from encoder-based methods like CustomNet, we introduce mechanisms specifically designed for multi-subject scenarios, aiming for harmonious and controllable generation without per-subject fine-tuning. 

\subsection{Controllable Generation and Datasets}

Beyond subject identity, precise control over generated images is crucial. ControlNet \cite{zhang2023adding} demonstrated powerful spatial control using conditioning inputs like depth maps, edges, or poses. Viewpoint control, as highlighted by Zero-1-to-3 \cite{liu2023zero} and utilized in CustomNet \cite{yuan2024customnet}, is vital for integrating subjects realistically into scenes. MS-CustomNet incorporates location control for multiple subjects via a reference image.

A significant bottleneck for training robust multi-subject models is the lack of suitable, publicly available datasets. Large-scale datasets like LAION \cite{schuhmann2022laion} are primarily image-text pairs, often lacking explicit multi-subject annotations or interaction details. Subject-centric datasets like COCO \cite{lin2014coco} provide bounding boxes and segmentations but may not capture the diversity needed for complex interaction learning in customized generation. 3D datasets like Objaverse \cite{deitke2023objaverse} are valuable for viewpoint learning but lack complex backgrounds and interactions. While some powerful multi-subject models have emerged, their training often relies on private or unreleased datasets, underscoring the importance of community efforts. The development of specialized, open datasets, such as the MSI dataset proposed in this work, derived from COCO with a focus on multi-subject interactions and structured inputs, is essential for training models like MS-CustomNet capable of handling complex compositional customization tasks.

\begin{figure*}[htbp]
    \centering
    \includegraphics[width=0.9\textwidth]{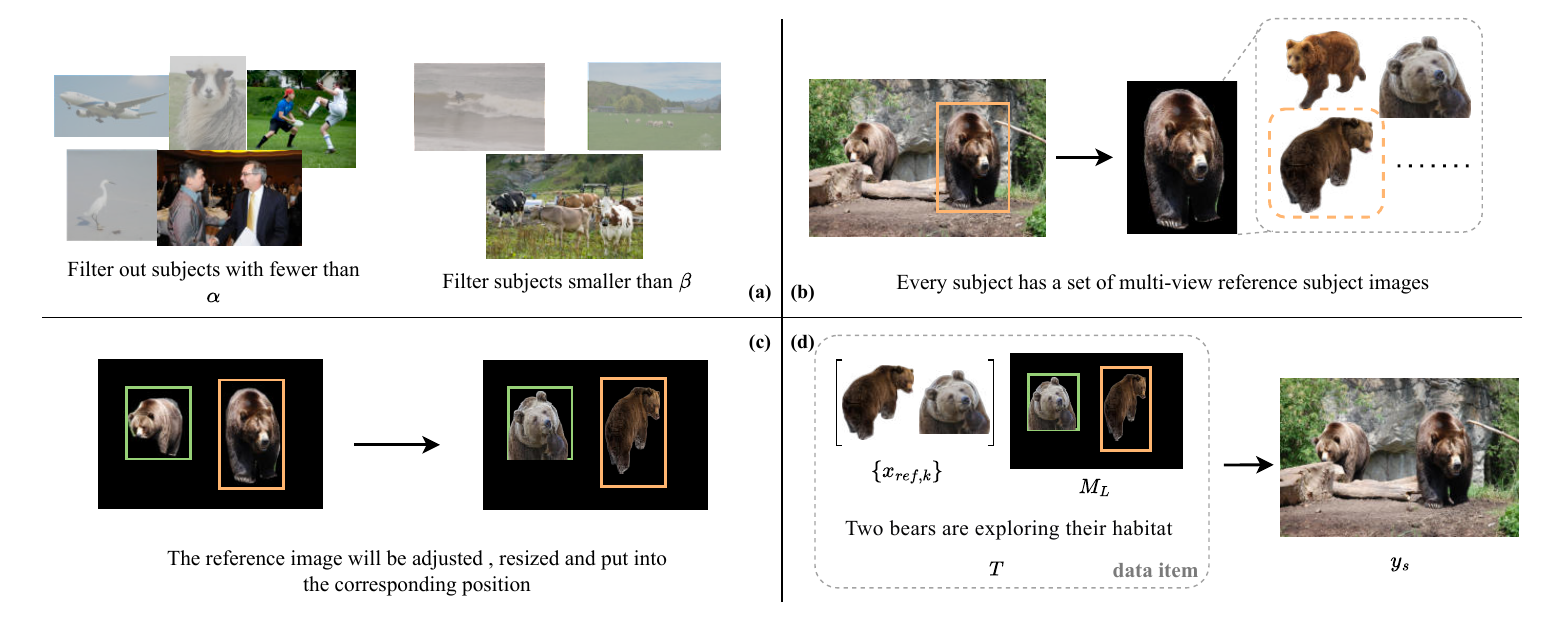}
    \caption{Construction pipeline of the MSI dataset. Key stages include COCO image filtering based on subject count and size, establishment of a multi-view reference subject pool, generation of composite layout maps using COCO annotations, and assembly of final training tuples.}
    \label{fig:dataset}
\end{figure*}

\section{Method}
\label{sec:method}

This section presents MS-CustomNet, a framework designed for the controllable generation of multiple subjects within a unified scene. Our approach integrates a novel dataset construction methodology with specific architectural enhancements to a diffusion model, enabling precise control over subject identities, their compositional structure and spatial relationship, all guided by textual descriptions.

\subsection{MSI Dataset Construction}

To effectively train a model for nuanced multi-subject customization, a dataset rich in examples of multiple subjects interacting within diverse scenes, complete with corresponding textual descriptions and subject-specific reference imagery, is essential. Addressing the scarcity of such datasets, we developed the MSI dataset, leveraging the COCO dataset \cite{lin2014coco}. The construction pipeline, illustrated in Fig. \ref{fig:dataset}, proceeds through several stages.

Initially, the COCO dataset undergoes a filtering process. A scene image, denoted as $y_{\text{s}}$, is selected based on the characteristics of its constituent subject instances. First, an individual subject instance $s_k$ within $y_{\text{s}}$ is identified as a candidate for consideration if its relative area satisfies the criterion:
\begin{equation}
\frac{\text{Area}(s_k)}{\text{Area}(y_{\text{s}})} > \beta,
\label{eq:subject_area_criterion}
\end{equation}
where $\text{Area}(\cdot)$ denotes the area of the subject's segmentation mask, and $\beta$ is a predefined relative area threshold for an individual subject's prominence. In our experiments, this was set to \(\beta = 0.015\), meaning a subject's segmentation mask must cover at least 1.5\% of the total image area to be considered salient.

Let $N(y_{\text{s}})$ represent the number of subject instances $s_k$ in image $y_{\text{s}}$ for which the condition in Eq. \eqref{eq:subject_area_criterion} holds. The image $y_{\text{s}}$ is retained if:
\begin{equation}
N(y_{\text{s}}) \ge \alpha,
\label{eq:image_selection_criterion}
\end{equation}
where $\alpha$ is a predefined minimum threshold for the number of qualifying subjects within the image. This threshold was set to \(\alpha = 2\). This dual stage filtering logic, depicted in Fig. \ref{fig:dataset} (a), ensures that chosen images feature a sufficient number of salient subjects, thereby providing rich contexts for interaction.

Subsequently, to mitigate trivial copy-pasting and promote model generalization, a pool of diverse reference images, $\mathcal{X}_{c_k} = \{ x_{\text{ref},c_k}^{(i)} \}_{i=1}^{M_{c_k}}$, is curated for each subject category $c_k$ identified in the filtered COCO subset. These pools contain $M_{c_k}$ varied instances of objects belonging to category $c_k$, as shown in Fig. \ref{fig:dataset} (b). During the training data generation, for a subject instance $s_k$ of category $c_k$ within the target image $y_{\text{s}}$, a reference image is randomly sampled:
\begin{equation}
x_{\text{ref},k} \sim \text{Uniform}(\mathcal{X}_{c_k}),
\label{eq:reference_sampling}
\end{equation}
This sampled image $x_{\text{ref},k}$ serves as the visual exemplar for the $k$-th subject in the scene.

The generation of a layout map, denoted as $M_L$, is then performed to enable explicit control over the compositional structure and spatial relationship of subjects. For each subject $s_k$ in $y_{\text{s}}$, its corresponding bounding box annotation $b_k$ and segmentation mask are obtained from COCO. The sampled reference subject image $x_{\text{ref},k}$ is resized and placed onto an initialized canvas, $M_0$, which is a matrix of zeros with dimensions corresponding to the target image size. This operation is iterated for all $K=N(y_s)$ selected salient subjects within the image, yielding the composite layout map $M_L$ as shown in Fig. \ref{fig:dataset} (c). This composition can be formally expressed as:
\begin{equation}
M_L = \mathcal{C}(\{(x_{\text{ref},k}, b_k, o_k)\}_{k=1}^{K}; M_0),
\end{equation}
where the notation $\mathcal{C}(\cdot)$ represents the compositing operation, signifying that the set of subjects $\{(x_{\text{ref},k}, b_k, o_k)\}_{k=1}^{K}$ are placed onto the initial canvas $M_0$. The term $o_k$ signifies the compositing order for subject $s_k$. This ordering determines the hierarchical relationship, or layering, among the subjects. To automate the construction of training instances, we impose a partial ordering on candidate objects by exploiting an occlusion prior derived from natural scene statistics: entities whose segmentation masks exhibit smaller pixel counts in the original COCO annotations are assigned a higher depth rank, thereby ensuring that spatially compact objects are consistently rendered as foreground layers occluding larger, background-dominant subjects. Consequently, it provides a mechanism for user-controlled specification of the compositional structure and spatial relationships within the final generated scene.

Finally, the MSI dataset is assembled, comprising tuples structured as:
\begin{equation}
\mathcal{D}_{\text{MSI}} = \left\{ \left( \{x_{\text{ref},k}\}_{k=1}^{K}, \mathcal{T}, M_L, y_{\text{s}} \right)_j \right\}_{j=1}^{|\mathcal{D}_{\text{MSI}}|},
\label{eq:msi_tuple_revised}
\end{equation}
Each tuple consists of the set of $K$ sampled reference subject images $\{x_{\text{ref},k}\}_{k=1}^{K}$, the textual prompt $\mathcal{T}$ which derived from the original COCO caption, the generated layout map $M_L$ specifying the compositional structure and spatial relationship, and the original filtered COCO image $y_{\text{s}}$ (Fig. \ref{fig:dataset} (d)).

The resulting MSI dataset constitutes a substantial and specialized resource for training and evaluating multi-subject customization models. Beginning with the complete set of 118,287 images from the COCO 2017 training partition, we implemented our stringent filtering protocol, governed by the criteria in Eq. \eqref{eq:subject_area_criterion} and \eqref{eq:image_selection_criterion}, with the subject count and area thresholds empirically set to \(\alpha=5\) and \(\beta=0.015\), respectively. This process yielded a curated collection of 14,537 scene images rich in compositional complexity. From these selected scenes, we constructed the final dataset, \(\mathcal{D}_{\text{MSI}}\). The dataset spans the full 80 object categories of COCO. To support robust identity preservation, we established a reference subject pool containing 2,400 exemplar images, meticulously collating 30 distinct instances for each category. Crucially, to preserve authentic aspect ratios and contextual details, all images within the MSI dataset, including target scenes \(y_s\), reference images \(x_{\text{ref},k}\), and their corresponding layout maps \(M_L\), retain their original height and width, without undergoing resizing or cropping. This meticulous construction ensures that \(\mathcal{D}_{\text{MSI}}\) provides the necessary structural and semantic diversity for training and rigorously evaluating models like MS-CustomNet on complex, controllable multi-subject composition tasks.

\subsection{MS-CustomNet Framework Architecture}

\begin{figure*}[htbp]
    \centering
    \includegraphics[width=0.9\textwidth]{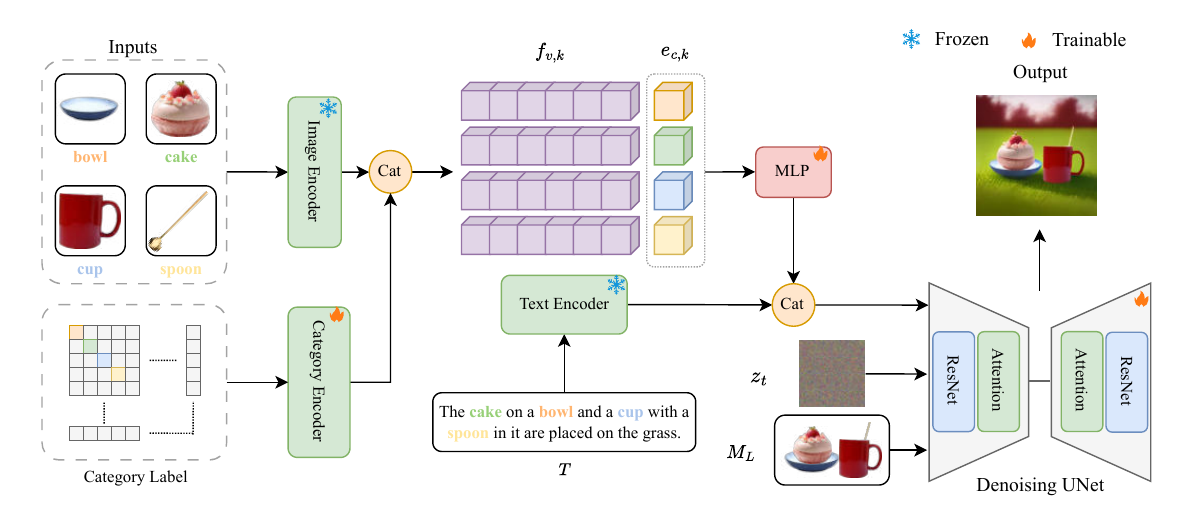}
    \caption{Architecture of MS-CustomNet. The model processes reference subject images (\(\{x_{ref,k}\}^K_{k=1}\)), textual prompts (\(\mathcal{T}\)), and layout maps (\(M_L\)). Category-aware subject features (\(f_{s,k}\)) are generated and, along with text and location cues, condition a latent diffusion model to synthesize the customized image.}
    \label{fig:backbone}
\end{figure*}

The MS-CustomNet framework is established upon a Latent Diffusion Model (LDM) \cite{rombach2022high}, as shown in Fig. \ref{fig:backbone}. Our implementation is based on the public CustomNet checkpoint. Its conditioning mechanisms are specifically augmented to simultaneously process multiple subject inputs, textual prompts, and explicit guidance for compositional structure and spatial relationship.

A key challenge is the association of textual descriptions with their corresponding visual reference inputs $\{x_{\text{ref},k}\}$. We address this by incorporating explicit category information. Let $c_k$ denote the category label for the $k$-th reference subject $x_{\text{ref},k}$. A category encoder $\Phi_c$ maps a one-hot encoded representation of $c_k$ to a category conditioning vector $\mathbf{e}_{c,k}$:
\begin{equation}
\mathbf{e}_{c,k} = \Phi_c(\text{one-hot}(c_k)),
\end{equation}

Visual features $\mathbf{f}_{v,k}$ are extracted from each reference image $x_{\text{ref},k}$ using a pre-trained CLIP \cite{radford2021clip} image encoder $\Phi_{\text{img}}$:
\begin{equation}
\mathbf{f}_{v,k} = \Phi_{\text{img}}(x_{\text{ref},k}),
\end{equation}

The category vector $\mathbf{e}_{c,k}$ is then concatenated with the visual features $\mathbf{f}_{v,k}$. This combined representation is projected by our newly introduced category-aware projection network $F_{\text{proj}}$ to produce the category-aware subject embedding $\mathbf{f}_{s,k}$:
\begin{equation}
\mathbf{f}_{s,k} = F_{\text{proj}}([\mathbf{f}_{v,k} ; \mathbf{e}_{c,k}]),
\end{equation}
where $[\cdot ; \cdot]$ denotes the concatenation operation. The set of these embeddings, $\mathcal{F}_s = \{\mathbf{f}_{s,k}\}_{k=1}^{K}$, serves as a conditional input to the U-Net of the diffusion model, integrated via cross-attention mechanisms.

Guidance for the compositional structure and spatial relationship is provided by the layout map $M_L$, generated as detailed in the above. $M_L$ encodes the desired 2D positions, scales, and relative layering of subjects, and can be manually adjusted by the user during inference. This map is encoded into a conditioning tensor $\mathbf{c}_L$ by an AutoencoderKL encoder $\Phi_L$:
\begin{equation}
\mathbf{c}_L = \Phi_L(M_L),
\label{eq:location_encoding_revised}
\end{equation}
The tensor $\mathbf{c}_L$ is usually concatenated channel-wise to the noisy latent variable $\mathbf{z}_t$ at each denoising step within the U-Net.

Textual conditioning is derived from the input prompt $\mathcal{T}$. This prompt is processed by a pre-trained CLIP text encoder $\Phi_{\text{t}}$ to yield text embeddings $\mathbf{f}_t$:
\begin{equation}
\mathbf{f}_t = \Phi_{\text{t}}(\mathcal{T}),
\label{eq:text_embedding_revised}
\end{equation}
These embeddings $\mathbf{f}_t$ are also fed into the U-Net,  modulating its intermediate layers through cross-attention.

The core generative mechanism utilizes a U-Net, denoted as $\epsilon_{\theta}$ and parameterized by $\theta$. This U-Net is tasked with predicting the noise component $\epsilon$ present in the noisy latent variable $\mathbf{z}_t$ at a given diffusion timestep $t$:
\begin{equation}
\hat{\epsilon} = \epsilon_{\theta}(\mathbf{z}_t, t, \boldsymbol{\phi}(t), \mathcal{F}_s, \mathbf{f}_t, \mathbf{c}_L),
\label{eq:unet_prediction}
\end{equation}
where $\boldsymbol{\phi}(t)$ represents the timestep embedding. The subject embeddings $\mathcal{F}_s$ and text embeddings $\mathbf{f}_t$ generally influence the U-Net's processing via distinct cross-attention blocks, while $\mathbf{c}_L$ is incorporated as described earlier. During the training process, the parameters of the pre-trained VAE and the CLIP encoder are kept frozen. Optimization is focused exclusively on the parameters of the U-Net and our projection network \(F_{\text{proj}}\).

\subsection{Training Strategies}

The training of MS-CustomNet focuses on optimizing its parameters to minimize the discrepancy between the predicted noise and the actual noise used to corrupt the latent representation of the target image.

The model parameters $\theta$ are optimized by minimizing the standard LDM loss function:
\begin{equation}
\mathcal{L}(\theta) = \mathbb{E}_{\substack{(y_{\text{s}}, \{x_{\text{ref},k}\}, \mathcal{T}, M_L) \sim \mathcal{D}_{\text{MSI}}, \\ \epsilon \sim \mathcal{N}(\mathbf{0},\mathbf{I}), t \sim \mathcal{U}(1,T)}}
\left\| \epsilon - \hat{\epsilon} \right\|^2_2,
\label{eq:loss_function_revised}
\end{equation}

The noisy latent $\mathbf{z}_t$ is derived from the target image $y_{\text{s}}$. First, $y_{\text{s}}$ is encoded into a clean latent $\mathbf{z}_0$ by a pre-trained VAE encoder $\Phi_{\text{l}}$:
\begin{equation}
\mathbf{z}_0 = \Phi_{\text{l}}(y_{\text{s}}),
\label{eq:vae_encode}
\end{equation}
Then, $\mathbf{z}_t$ is obtained through the forward diffusion process:
\begin{equation}
\mathbf{z}_t = \sqrt{\bar{\alpha}_t} \mathbf{z}_0 + \sqrt{1-\bar{\alpha}_t} \epsilon,
\label{eq:noising_process}
\end{equation}
where $\bar{\alpha}_t$ is the noise schedule parameter at timestep $t$, $\epsilon$ is sampled from a standard normal distribution, and $T$ is the total number of diffusion timesteps. The conditioning inputs $\mathcal{F}_s$, $\mathbf{f}_t$, and $\mathbf{c}_L$ are derived as previously described.

The Dual Stage Training (DST) regimen is adopted to structure the learning process. The first stage focuses on layout and interaction priming. During this stage, the model is trained with a learning rate $\eta_1$ for $E_1$ epochs, emphasizing accurate subject placement according to $M_L$ and the learning of basic inter-subject interactions as guided by $\mathcal{T}$. The second stage is dedicated to identity and detail refinement. After $E_1$ epochs, the learning rate is reduced to $\eta_2$, and training continues for an additional $E_2$ epochs, with $E_1 > E_2$. This stage prioritizes the refinement of subject-specific details, enhancing identity fidelity and overall photorealism. The transition between stages occurs when the current epoch $e$ exceeds $E_1$.

Furthermore, a Curriculum Learning on Subject Quantity (CLSQ) strategy is employed. This involves progressively increasing the maximum number of subjects, $K(e)$, permitted in a training sample as training progresses through epoch $e$. Let $K_{\text{min}}$ and $K_{\text{max}}$ be the minimum and maximum number of subjects considered, respectively, and $E$ be the total number of training epochs. The number of subjects for epoch $e$, $K(e)$, is determined by:
\begin{equation}
K(e) = \max \left( K_{\text{min}}, \, K_{\text{min}} + \left\lfloor \left( \frac{e}{E} \right)^{\gamma} \cdot (K_{\text{max}} - K_{\text{min}}) \right\rfloor \right),
\end{equation}
where $\lfloor \cdot \rfloor$ denotes the floor function, $E$ is the total number of training epochs, and $\gamma$ is a parameter controlling the pace of curriculum progression. A smaller $\gamma$ leads to a faster initial increase in $K(e)$, while a larger $\gamma$ results in a more gradual start. 

If a training sample originally contains $K$ subjects, where $K > K(e)$, a random subset of $K(e)$ subjects is selected from the original set $\mathcal{S}_{\text{ori}} = \{s_1, \dots, s_{K}\}$:
\begin{equation}
\mathcal{S}(e) = \text{RandomSubset}(\mathcal{S}_{\text{ori}}, \min(K(e), K)),
\end{equation}
Otherwise, if $K \le K(e)$, all $K$ subjects are used. This strategy allows the model to initially learn simpler compositions before gradually adapting to more complex multi-subject scenarios.

\begin{figure*}[hbtp]
\centering
\includegraphics[width=0.9\textwidth]{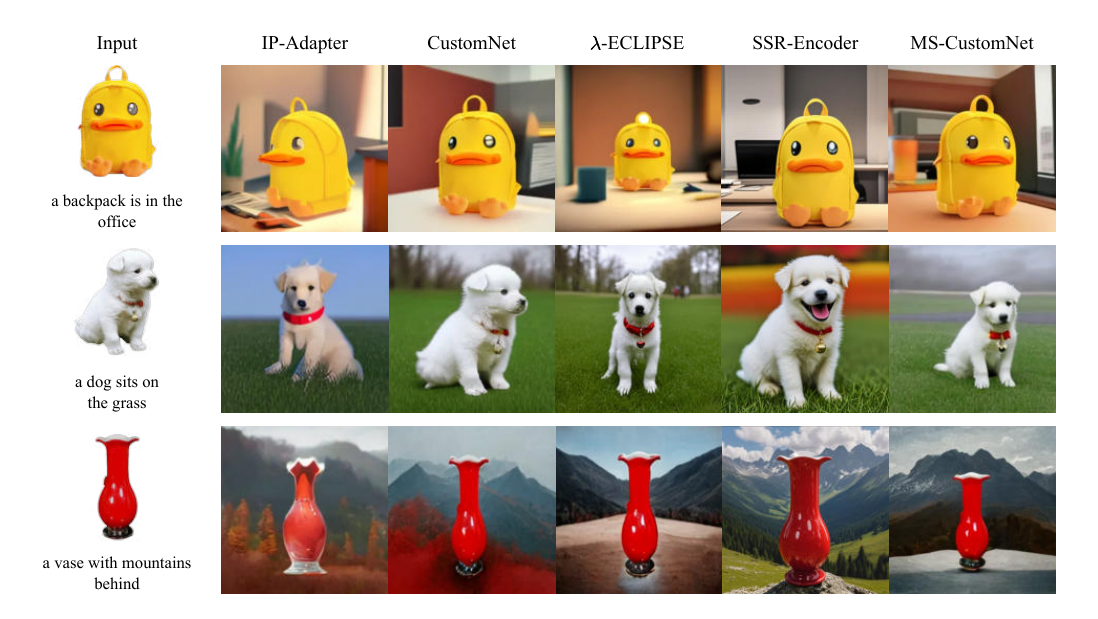}
\caption{Qualitative assessment of single-subject customization. The figure showcases the MS-CustomNet's proficiency in preserving subject identity and integrating subjects cohesively into diverse scenes, guided by varied textual prompts.}
\label{fig:ss}
\end{figure*}

\section{Experiments}

This section details the experimental setup, evaluation protocols, and comparative analyses conducted to assess the performance of MS-CustomNet. We evaluate its capabilities in both single-subject and multi-subject customization tasks, with a particular focus on its ability to preserve subject identity, adhere to textual prompts, and manage complex compositional structures.

State-of-the-art methods such as MS-Diffusion are excluded from direct comparison in this study because they rely on large-scale proprietary datasets that are not publicly available and employ substantially larger model architectures. In contrast, our proposed lightweight model, MS-CustomNet, is specifically designed for efficient deployment and to offer users explicit control over multi-object scenes. Rather than aiming to match the raw image-generation fidelity of SOTA methods, MS-CustomNet focuses on user-defined spatial relationships. Accordingly, our experimental evaluation centers on quantifying MS-CustomNet’s ability to realize precise, user-controlled spatial arrangements.

\subsection{Experimental Setup}

Our implementation is built upon the publicly available checkpoint of CustomNet, with all experiments conducted on a server equipped with two NVIDIA V100 GPUs, each providing 32GB of VRAM. We employ the AdamW optimizer for model training. The model is trained using a Dual Stage Training regimen for a total of 10 epochs, which consists of an initial 7-epoch stage with a learning rate of \(1 \times 10^{-4}\), followed by a 3-epoch refinement stage with a reduced learning rate of \(5 \times 10^{-5}\). Furthermore, for our Curriculum Learning by Subject Quantity strategy, the number of subjects is progressively increased from a minimum of \(K_{\text{min}} = 2\) to a maximum of \(K_{\text{max}} = 5\), governed by a curriculum pace of \(\gamma = 1.0\). To stabilize optimization we train with automatic mixed precision in FP16, achieving an effective batch size of 4.

To rigorously evaluate multi-subject customized image generation, we constructed a dedicated benchmark named MSIBench. For each subject category within the COCO dataset, a pool of ten representative subject exemplars was curated by sourcing images from the web. In each evaluation instance, between two and five images are randomly selected from this pool to serve as the customized subjects. Subsequently, GPT was employed to generate a textual description outlining the desired background context for these subjects. This process yielded a benchmark collection of 100 multi-subject customization tasks, each comprising reference subject images and a corresponding textual prompt.

Consistent with established practices in generative model evaluation, we employ several quantitative metrics to assess distinct aspects of generation quality. The fidelity of customized subjects in the generated image to their original references, or identity preservation, is quantified using feature similarity. For this, we utilize DINO \cite{zhang2022dino} features, referred to as DINO-I, and CLIP \cite{radford2021clip} image features, denoted as CLIP-I. Higher scores for both DINO-I and CLIP-I indicate better identity preservation. The semantic congruence between the textual background prompt and the visual content of the generated image's background is measured using CLIP text-image similarity, termed CLIP-B, where higher scores denote better alignment.

For assessing the accuracy of subject rendering specifically in multi-subject scenarios, we use a pre-trained YOLO detector \cite{khanam2024yolov11}. This involves two metrics: YOLO-Subj evaluates the consistency between the number and categories of subjects depicted in the generated image and those specified in the input. Concurrently, YOLO-L measures the positional accuracy of generated subjects relative to user-specified target locations in $M_L$, often using Intersection over Union (IoU) between detected boxes and target locations, with higher scores indicating better localization. In our evaluations, single-subject customization relies on DINO-I, CLIP-I, and CLIP-B, while for multi-subject customization, all aforementioned metrics, including YOLO-Subj and YOLO-L, are employed.

For single-subject customization, evaluation relies on DINO-I, CLIP-I, and CLIP-B. For multi-subject customization, all metrics, including YOLO-Subj and YOLO-L, are employed.

\begin{table*}[htbp]
    \centering
    \caption{Quantitative comparison for single-subject and multi-subject customization.}
    \label{tab:ss-ms}
    \resizebox{0.9\textwidth}{!}{
    \begin{tabular}{l|ccc|ccccc}
        \toprule
        \multirow{2}{*}{Method} & \multicolumn{3}{c|}{Single-Subject Customization} & \multicolumn{5}{c}{Multi-Subject Customization} \\
        & DINO-I $\uparrow$ & CLIP-I $\uparrow$ & CLIP-B $\uparrow$ & YOLO-Subj $\uparrow$ & YOLO-L $\uparrow$ & DINO-I $\uparrow$ & CLIP-I $\uparrow$ & CLIP-B $\uparrow$ \\
        \midrule
        IP-Adapter      & 0.58 & 0.75 & 0.29 & --- & --- & --- & --- & --- \\
        CustomNet       & \textbf{0.77} & \textbf{0.82} & 0.33 & --- & --- & --- & --- & --- \\
        $\lambda$-ECLIPSE & 0.61 & 0.78 & 0.34 & 0.62 & --- & 0.56 & 0.77 & 0.30 \\
        SSR-Encoder     & 0.67 & 0.79 & \textbf{0.36} & \textbf{0.71} & 0.91 & 0.57 & 0.79 & \textbf{0.35} \\
        MS-CustomNet    & 0.74 & \textbf{0.82} & 0.32 & 0.68 & \textbf{0.94} & \textbf{0.61} & \textbf{0.79} & 0.34 \\
        \bottomrule
    \end{tabular}
    }
    \vspace{0.5em}
\end{table*}

\begin{figure*}[htbp]
    \centering
    \includegraphics[width=0.9\textwidth]{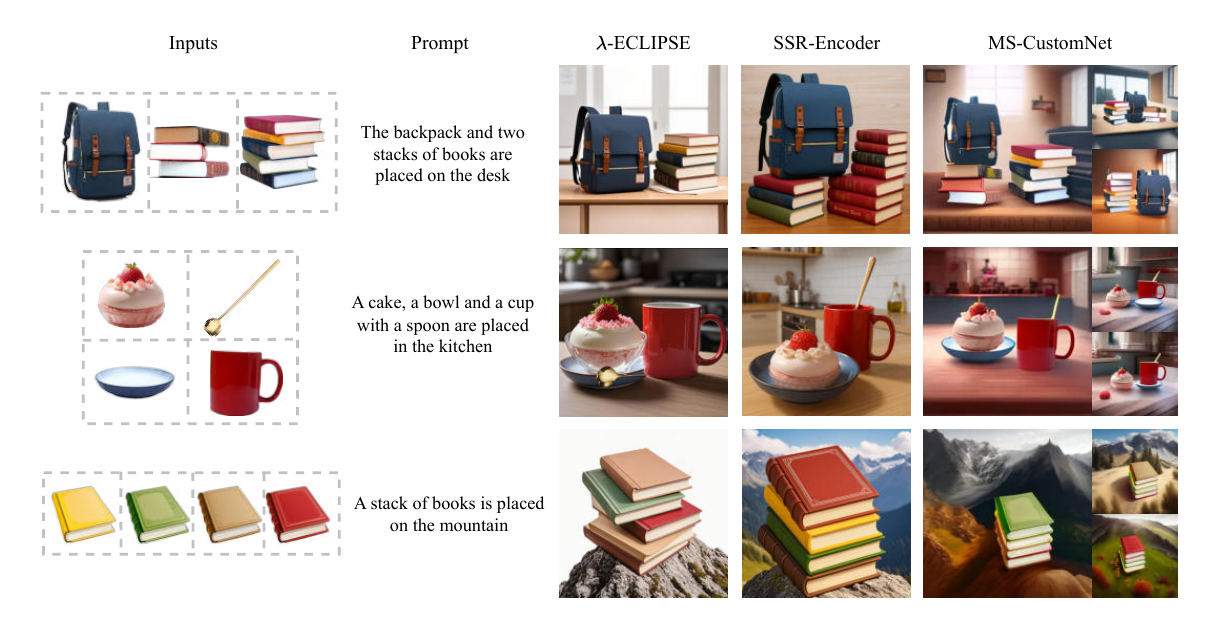}
    \caption{Qualitative assessment of multi-subject customization. These examples illustrate the MS-CustomNet's capacity to generate intricate scenes featuring multiple customized subjects, effectively preserving individual identities while adhering to user-defined compositional and spatial relationship.}
    \label{fig:ms}
\end{figure*}

\subsection{Comparative Analysis}

We conduct a comprehensive comparative analysis of MS-CustomNet against state-of-the-art methods in both single-subject and multi-subject customization. Our evaluation focuses on benchmarking MS-CustomNet against prominent existing approaches, namely IP-Adapter \cite{ye2023ip}, CustomNet \cite{yuan2024customnet}, $\lambda$-ECLIPSE \cite{patel2024eclipse}, and SSR-Encoder \cite{zhang2024ssr}. While current state-of-the-art (SOTA) methods like MS-Diffusion excel in general multi-subject image quality leveraging large-scale proprietary datasets, MS-CustomNet's primary objective is distinct: we prioritize providing explicit, fine-grained user control over the compositional structure and spatial relationships of multiple subjects within a more accessible and efficient framework. Given these fundamentally different focuses, a direct comparison of overall generation quality with such large-scale SOTA methods is not undertaken in this work.

\subsubsection{Single-Subject Customization}
MS-CustomNet is benchmarked against the aforementioned leading approaches for single-subject customization tasks. Conceptually extending the robust foundation of CustomNet, MS-CustomNet demonstrates highly competitive performance in generating single customized subjects. As illustrated qualitatively in Fig. \ref{fig:ss} and quantitatively in Table \ref{tab:ss-ms}, MS-CustomNet effectively preserves the salient visual characteristics and identity of the input subject. For instance, in the examples provided, the distinctive features of the yellow duck backpack, the white puppy, and the red vase are faithfully replicated.

A key characteristic of MS-CustomNet in this context is its ability to not only maintain high fidelity to the subject's distinguishing features but also to achieve a commendable degree of adaptive viewpoint generalization. This allows the subject to be rendered from varied perspectives, ensuring it is well-integrated into the novel scene specified by the textual prompt while maintaining coherence and naturalism. For example, the puppy, originally shown from a side profile, is convincingly rendered from slightly different angles in the generated images, adapting to the grassy environment. Similarly, the backpack and vase are seamlessly placed into office and mountainous backdrops, respectively, with appropriate lighting and perspective.

Comparing its performance, MS-CustomNet achieves DINO-I and CLIP-I scores that are highly competitive with, and in some aspects exceed, those of specialized single-subject methods. For instance, while IP-Adapter can sometimes exhibit a loss of fine-grained subject details or introduce undesirable alterations, MS-CustomNet offers more consistent feature retention. $\lambda$-ECLIPSE provides strong customization but can occasionally introduce subtle stylistic shifts or minor identity drift. SSR-Encoder excels in producing highly photorealistic outputs and strong identity preservation. MS-CustomNet, while also achieving high identity scores, builds upon the principles of CustomNet to offer a robust alternative, demonstrating comparable, if not superior in some evaluations, identity preservation and scene integration. The results affirm its parity with, and often favorable performance against, established methods in single-subject generation tasks, providing a solid foundation for its extension to more complex multi-subject scenarios where fine-grained control is paramount. Furthermore, its ability to accurately interpret prompt semantics ensures that the generated background aligns well with the user's textual description, leading to harmonious and contextually appropriate compositions.

\subsubsection{Multi-Subject Customization}

We focus on a direct comparison of the performance of MS-CustomNet with $\lambda$-ECLIPSE and SSR-Encoder, particularly in terms of controllable spatial structure metrics, aiming to quantify the advantages and disadvantages of each method regarding user-specified hierarchy and spatial control. MS-CustomNet distinguishes itself by extending single-subject customization paradigms to the more complex multi-subject domain, crucially empowering users with explicit, fine-grained control over how subjects are arranged and interact compositionally. This capability is pivotal for generating scenes that adhere to precise, user-defined hierarchical and spatial configurations.

For instance, MS-CustomNet facilitates the explicit definition of compositional structure, such as the stacking order in a pile of books, allowing users to dictate precisely which book appears on top of another, thereby controlling mutual occlusion and relative depth. Similarly, in a scenario involving a cake and a bowl, our framework enables users to specify nuanced spatial relationships: one can direct the model to render the cake inside the bowl, implying a specific layering where the bowl partially contains or is positioned behind the cake from a compositional perspective. Alternatively, the user can stipulate that the cake is placed behind the bowl, altering the occlusion and perceived depth relationship. This explicit control over the z-ordering and containment semantics is a key differentiator.

In contrast, while methods like SSR-Encoder employ masks to guide subject localization, this approach primarily dictates the 2D placement of subjects within the image plane. Such mask-based conditioning, though effective for general positioning, does not readily extend to the explicit specification of the intricate compositional hierarchy or the precise spatial interplay that MS-CustomNet is designed to manage. Consequently, SSR-Encoder may struggle to disambiguate or enact user intentions for complex layered arrangements like the aforementioned book stack or the cake-in-bowl versus cake-behind-bowl scenarios, as its control mechanism lacks the granularity to define these hierarchical relationships explicitly.

As demonstrated in Fig. \ref{fig:ms} and Table \ref{tab:ss-ms}, MS-CustomNet not only preserves subject identities but also excels in adhering to these user-specified compositional structures and spatial relationships. The high YOLO-L score quantitatively supports its proficiency in precise localization, which, in conjunction with its architectural design for compositional control, enables the generation of semantically coherent and structurally accurate multi-subject images. In scenarios where precise control over synthetic structures and spatial hierarchies is paramount, MS-CustomNet offers compelling advantages. Furthermore, its framework achieves this enhanced controllability more efficiently than multi-subject generation approaches that lack such explicit control mechanisms and may require more complex inference-time manipulations or significantly larger models to implicitly learn these relationships.

\begin{figure}[htbp]
\centering
\includegraphics[width=0.45\textwidth]{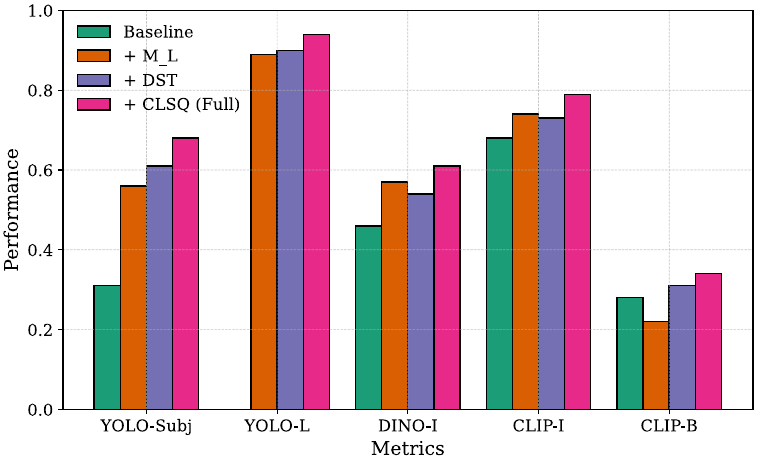}
\caption{Bar chart comparing various metrics under different ablation configurations. The horizontal axis shows the four configurations in order: baseline, introduction of \(M_L\), introduction of DST, and introduction of CLSQ.}
\label{fig:ab}
\end{figure}

\subsection{Ablation Study}

\begin{figure*}[htbp]
\centering
\includegraphics[width=0.9\textwidth]{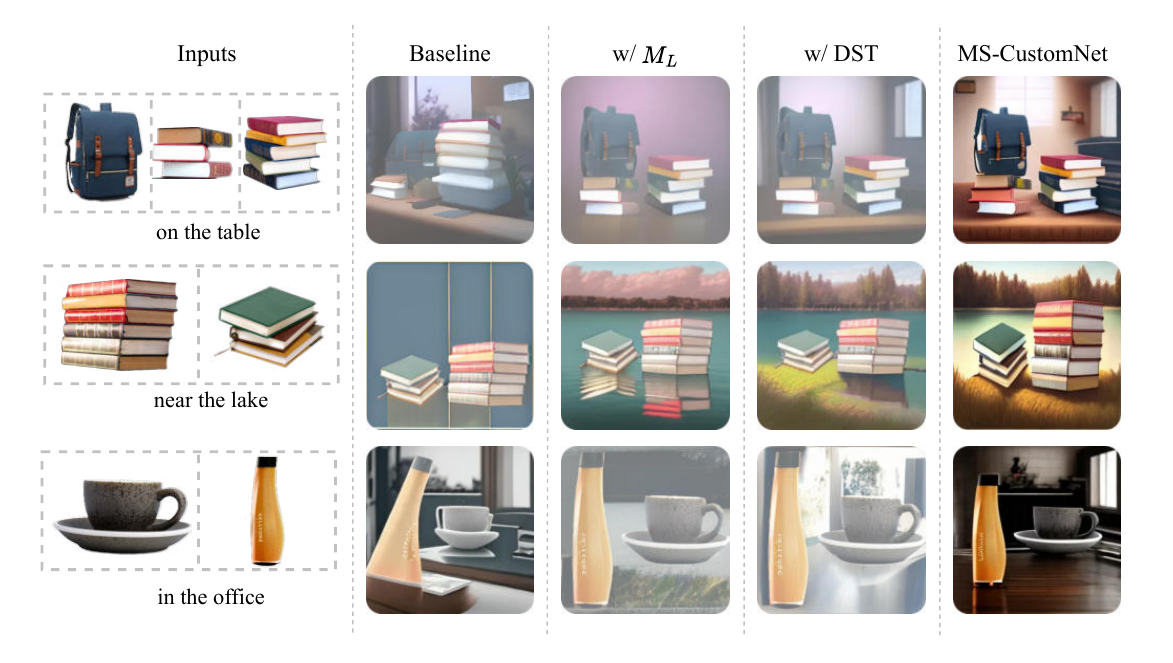}
\caption{Visual progression in the ablation study of MS-CustomNet. This figure illustrates the stepwise enhancement in generated image quality, showing improvements in identity preservation, subject clarity, compositional accuracy, and background coherence as components are added, from the Baseline to the full MS-CustomNet model incorporating curriculum learning.}
\label{fig:compare}
\end{figure*}

To evaluate our contributions, we performed an ablation study by incrementally adding each key component—$M_L$ guidance, DST, and CLSQ—to a baseline model. The resulting quantitative metrics (Fig. \ref{fig:ab}) and qualitative outputs (Fig. \ref{fig:compare}) reveal the stepwise impact of each innovation.

\subsubsection{Baseline}
Our investigation commences with a baseline model, which consists of the pre-trained diffusion model fine-tuned on the MSI dataset without the integration of our proposed mechanisms. Specifically, this configuration does not utilize the layout map $M_L$ for spatial guidance. The performance of this rudimentary model, as quantified in Figure \ref{fig:ab}, is predictably suboptimal. The YOLO-Subj score is exceptionally low, indicating a significant deficiency in adhering to the specified object count and categories from the input prompt. As the model lacks any explicit spatial guidance, the YOLO-L metric is not applicable at this stage. Furthermore, the identity preservation metrics, DINO-I and CLIP-I, register poor performance. This quantitative deficiency is corroborated by the visual outputs in Figure \ref{fig:compare}, where subjects exhibit severe deformation, feature corruption, and inconsistent appearances. Although the CLIP-B score, which measures text-background alignment, is comparatively moderate, the generated scenes often suffer from chaotic composition and a failure to establish a coherent background that aligns with the textual prompt, thereby compromising the overall semantic integrity and visual quality of the output.

\subsubsection{Introduction of Location Information}
Upon the integration of the layout map $M_L$ during the training phase, the model's generative capabilities undergo a foundational shift. As illustrated in Figure \ref{fig:ab}, this addition yields a dramatic improvement in object-centric metrics. The YOLO-Subj score increases substantially, and YOLO-L achieves a high value, collectively signifying a marked improvement in the model's capacity for spatial reasoning and compositional accuracy. The model now reliably generates the correct number of objects in their designated locations. A moderate increase in identity preservation, DINO-I and CLIP-I, is also observed, likely due to the improved object isolation preventing feature bleeding.

However, this advancement reveals a critical representational trade-off: the CLIP-B score experiences a sharp decline. This phenomenon suggests that by heavily conditioning the model on precise spatial layouts, the generative process over-prioritizes object placement at the expense of rendering a rich, contextually-aligned background. The qualitative results in Figure \ref{fig:compare} vividly illustrate this trade-off. While subjects are correctly positioned and more distinct, they appear pasted onto hazy, underdeveloped, or monochromatic backgrounds that lack semantic depth and detail. This results in images that are compositionally correct but stylistically disjointed and environmentally impoverished.

\subsubsection{Introduction of DST}
To mitigate the aforementioned limitations, particularly the degradation in subject fidelity and background coherence, we introduced the DST. The results in Figure \ref{fig:ab} demonstrate the efficacy of this component. The model maintains its high performance on YOLO-Subj and YOLO-L, confirming that the gains in spatial control are preserved. More importantly, we observe a notable enhancement in both identity preservation metrics of DINO-I and CLIP-I. This indicates that DST effectively refines the subject's textural and structural features, leading to higher fidelity and better preservation of unique visual characteristics. A modest recovery in the CLIP-B score is also observed, suggesting an improved alignment between the generated background and the textual prompt.

Despite these improvements, the qualitative examples in Figure \ref{fig:compare} reveal a lingering challenge. While the background begins to form more coherent structures, the model occasionally struggles with rendering high-frequency details across both the subject and the scene. In rendering subjects with intricate features, the outputs can exhibit localized blurring or a loss of textural sharpness, which tempers the overall visual impact and photorealism.

\subsubsection{Introduction of CLSQ}
The final component, CLSQ is introduced to orchestrate the complex learning process and resolve the competing objectives of subject fidelity, compositional accuracy, and background realism. The full MS-CustomNet architecture, equipped with CLSQ, demonstrates superior performance across all evaluated metrics, achieving a synergistic optimum. As depicted in Figure \ref{fig:ab}, the YOLO-Subj and YOLO-L metrics attain their peak values, signifying unparalleled accuracy in object-level semantic and spatial control. The identity fidelity metrics, DINO-I and CLIP-I, also reach their highest points, reflecting exceptional identity preservation without compromise.

Most critically, the CLIP-B metric exhibits a substantial improvement, surpassing all previous configurations. This underscores the model's newfound ability to synthesize complex, detailed, and semantically rich backgrounds that are highly consistent with the textual descriptions. The staged learning process of CLSQ enables the model to first master foundational aspects before progressing to more nuanced tasks, thus preventing the representational trade-offs observed in earlier stages. The qualitative results in Figure \ref{fig:compare} provide compelling visual evidence. The final images showcase a harmonious integration of high-fidelity subjects within coherent and contextually appropriate scenes. This demonstrates that CLSQ successfully guides the model to balance these multifaceted generation requirements, culminating in a marked enhancement of overall photorealism and semantic integrity.

\section{Conclusion}
This paper introduced MS-CustomNet, a novel method for multi-subject customization, alongside the specialized MSI dataset. MS-CustomNet empowers users to generate images featuring multiple customized subjects, preserving their identities while crucially offering explicit control over their compositional structure and spatial relationships within the scene. The MSI dataset facilitates training for such intricate scenarios. These contributions significantly advance user-directed, multi-subject image generation, particularly for applications demanding precise control over how multiple elements are composed and spatially arranged.

\bibliographystyle{IEEEtran}

\bibliography{main}

\end{document}